\documentclass[runningheads]{llncs}
\usepackage[misc]{ifsym}
\usepackage[table]{xcolor}
\usepackage{subfigure}
\usepackage{graphicx}
\usepackage{paralist}
\usepackage{color}
\usepackage{mfirstuc}
\usepackage{setspace}
\usepackage{graphicx}
\usepackage{etoolbox}
\usepackage[export]{adjustbox}
\usepackage{array}
\usepackage{float}
\usepackage{amsmath}
\usepackage{amssymb}
\usepackage{amsfonts}
\usepackage{cite}
\usepackage{color}
\usepackage{booktabs}
\usepackage{multicol,multirow}
\begin{document}
\title{Cross-Modal Attention Alignment Network with Auxiliary Text Description for zero-shot sketch-based image retrieval}
\titlerunning{Cross-Modal Attention Alignment Network with Auxiliary Text Description}
%
%

\author{Hanwen Su\orcidID{0009-0003-4335-5290} \and Ge Song \thanks{Corresponding author}\orcidID{0000-0002-2159-8203}  \and Kai Huang\orcidID{0009-0000-1735-1788} \and Jiyan Wang\orcidID{0009-0000-4962-4496} \and Ming Yang\orcidID{0000-0001-8936-4270}}

\authorrunning{Su. et al.}

\institute{School of Computer and Electronic Information, Nanjing Normal University, Nanjing, 210023, CHN}

\maketitle     

\begin{abstract}

In this paper, we study the problem of \textit{zero-shot sketch-based image retrieval} (ZS-SBIR). The prior methods tackle the problem in a two-modality setting with only category labels or even no textual information involved. However, the growing prevalence of Large-scale pre-trained Language Models (LLMs), which have demonstrated great knowledge learned from web-scale data, can provide us with an opportunity to conclude collective textual information. Our key innovation lies in the usage of text data as auxiliary information for images, thus leveraging the inherent zero-shot generalization ability that language offers. To this end, we propose an approach called Cross-Modal Attention Alignment Network with Auxiliary Text Description for zero-shot sketch-based image retrieval. The network consists of three components: (i) a Description Generation Module that generates textual descriptions for each training category by prompting an LLM with several interrogative sentences, (ii) a Feature Extraction Module that includes two ViTs for sketch and image data, a transformer for extracting tokens of sentences of each training category, finally (iii) a Cross-modal Alignment Module that exchanges the token features of both text-sketch and text-image using cross-attention mechanism, and align the tokens locally and globally. Extensive experiments on three benchmark datasets show our superior performances over the state-of-the-art ZS-SBIR methods.

\keywords{ Zero-shot Learning \and LLMs \and Sketch-based Image Retrieval.}
\end{abstract}
\section{Introduction}

\begin{figure}[h]
    \centering
    \subfigure[]
    {\includegraphics[width=8cm]{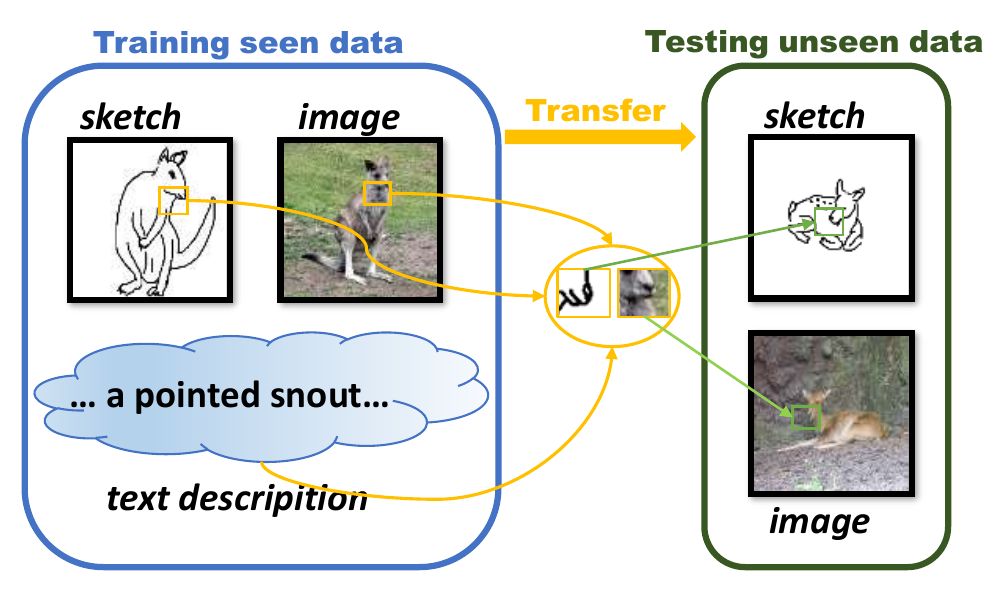}}
    \subfigure[]
    {\includegraphics[width=8cm]{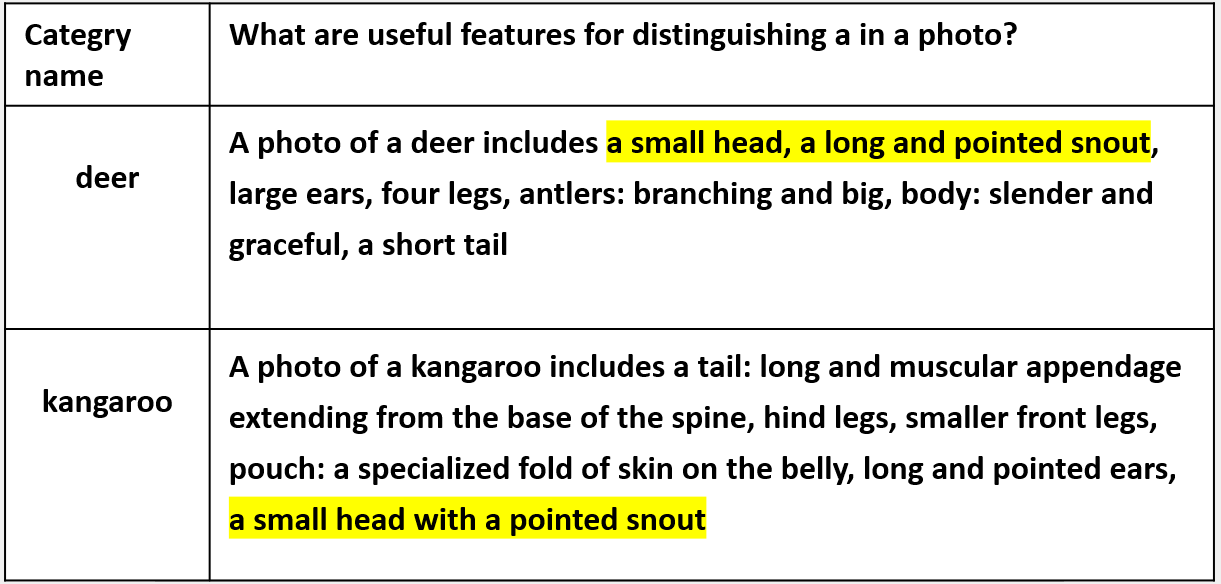}}
    \caption{(i) During training, the sketch, image, and the corresponding text are fed into the model to learn the alignment for the correspondence of a specific region. (ii) For inference, the transferred knowledge is utilized to do the ZS-SBIR.}
    \label{fig:intro}
\end{figure}

Sketch-based image retrieval (SBIR) is a practical problem that uses a hand-drawn sketch as the query to retrieve the image of interest in the gallery. There are many application scenarios for this problem, such as e-commerce: we can retrieve a pair of shoes we want to purchase by drawing a sketch. The conventional SBIR requires the training and testing data to come from the distributions of the same categories, which limits its real-world applications.  Moreover, the performances of traditional SBIR models are poor on the data of unseen classes. In addition, annotating large quantities of images is labor-intensive and time-consuming. Driven by practical constraints, late research on SBIR had shifted to zero-shot setup, i.e., zero-shot sketch image retrieval (ZS-SBIR)~\cite{ref_domain,ref_eff,ref_tvt,ref_sk3t,ref_semantic,ref_zse,ref_graph,ref_tied,ref_style,ref_doodle,ref_progress,ref_frame} because of the prevailing data-scarcity problem~\cite{ref_style,ref_frame,ref_fast}. The basic assumption behind the standard ZSL (zero-shot learning) methods is that test data only contain unknown classes\cite{ref_incre,ref_yang1}. ZS-SBIR is more challenging due to the cross-modal and zero-shot paradigm's inherent modal and semantic gap\cite{ref_yang2,ref_yang3}. So far, many methods have made progress on ZS-SBIR. While~\cite{ref_tvt,ref_sk3t,ref_zse} tried methods with no class labels involved,~\cite{ref_domain,ref_eff,ref_semantic,ref_graph,ref_tied,ref_style,ref_progress} only used class labels embeddings during training. However, recent works have shown that text documents from internet sources can provide great auxiliary information for ZSL~\cite{ref_i2v}. In this case, we try to make use of the text information to boost the performance of ZS-SBIR models. Therefore, our motivation can be summarized as: (i) Because images inherently include complex information, such as coloration and background, simply utilizing the alignment of sketch and image features can bring limited benefit to model's generalization ability. (ii) As shown in Fig.\ref{fig:intro}, since the detailed visual features of a sketch or image can be summarized in texts, we can align the regions between a sketch and an image with the help of the same phrases. In other words, textual information can be treated as the summary of both sketches and images. (iii) Besides, textual information itself contains very rich generalization knowledge, making it possible for us to use it as the bridge between sketch and image.

Thus, our main approach treating language as an auxiliary representation for ZS-SBIR, trying to boost the existing models¡¯ zero-shot generalization capability. Then, the problem then lies in how the textual information should be leveraged. However, how to get detailed and conclusive texts? Thanks to Large-scale pre-trained Language Models (LLMs), such as GPT-3~\cite{ref_lang}, which has shown brilliant world knowledge on a wide range of topics, we can collect texts in a k-prompt way. The prompt can be, for example, \textit{"A caption describing a photo of a \{category name\}."}~\cite{ref_cascade} or in other formats, which we will discuss in Section 4.3. We then collect and pick some reasonable outputs as our textual information for a certain category. Although generating and collecting the text information for each category requires human labor to some extent, it's more labor-saving than annotating the training images and sketches. The model then learns to align the regions between a sketch and an image. Thus learned knowledge can be generalized to the unseen categories to boost the performance of ZS-SBIR methods.

In this work, we introduce a new framework to handle the ZS-SBIR problem, utilizing the auxiliary textual information. The framework consists of three components: (i) a Description Generation Module that generates texts describing the distinguishing features for each training category by prompting an LLM with several interrogative sentences, and (ii) a Feature Extraction Module that includes two ViTs for both sketch and image data, a Transformer for extracting tokens of sentences of each training category, specifically, (iii) a Cross-modal Alignment Module that exchanges the modal-specific semantic information of both text-sketch and text-image using cross-modal attention, and align the tokens locally and globally. Extensive experiments on three benchmark datasets of ZS-SBIR verify the superiority of our method. We summarize the main contributions of this work as follows:
\begin{compactitem}
\item We introduce textual information into the framework of ZS-SBIR model. More specifically, we leverage texts to capture and align the sketches and images local features with more zero-shot generalization ability.

\item We design several prompts, and collect more thorough descriptions for training categories, equipped with a global triplet loss of three modalities and a local matching loss, to acquire knowledge for ZS-SBIR.

\item We conduct experiments on 3 benchmark datasets of ZS-SBIR, achieving a state-of-the-art result.
\end{compactitem}

\section{Related Work}
\subsection{Zero-shot Sketch-Based Image Retrieval}
ZS-SBIR aims at generalizing the knowledge learned from the seen training class to unseen testing categories using a query sketch to retrieve the images of the same category. The problem is introduced by~\cite{ref_frame}, which aims to reduce the domain gap between sketches and photos by an image-to-image translation approximation. The subsequent work such as~\cite{ref_domain,ref_eff,ref_semantic,ref_graph,ref_tied,ref_style,ref_progress} used the CNNs as backbones to extract features and then utilize projections to create a joint space. Despite the help of joint semantic space, ~\cite{ref_semantic} introduced the generative adversarial network for alignment. While, \cite{ref_tvt,ref_zse} used ViT\cite{ref_16} as its backbone, achieving better results compared to CNN-based methods. Moreover, the teacher-student paradigm was employed by\cite{ref_domain,ref_eff} to distill the knowledge. In addition, a test-time training paradigm\cite{ref_sk3t} on the sketches in the test set was introduced to adapt to the test set distribution. \cite{ref_zse} addressed the importance and the generalization capability of aligning the corresponding local patches for ZS-SBIR. However, the above-mentioned methods got sub-optimal results, because they didn't make use of the potential zero-shot generalization ability that textual information has.
\subsection{Cross Attention Based on Transformer}
The Transformer architecture was first introduced for machine learning by \cite{ref_att}. After that, \cite{ref_16} utilized the Vision Transformer (ViT) directly to the sequence of image patches, achieving great results in the image recognition task. Later methods\cite{ref_crossvit,ref_crosstr} adopted the ViT as their backbones and made use of the cross attention's reasoning capability. While \cite{ref_crosstr} finds the pixel-level correspondence for images in the few-shot image classification task, \cite{ref_crossvit} later offers a dual-branch ViT that can extract tokens at multiple scales and exchange class tokens between two branches, enhancing the class token for classification. Besides, for cross-modal retrieval task, \cite{ref_rethink} directly shares a unified multi-head attention classification network for the two modalities, and therefore make better use of the region structural information shared by the cross-modal instance itself. More recently, \cite{ref_tvt} introduced a shared fusion ViT that can offer an extra fusion token for interaction with other tokens through the self-attention mechanism, resulting in a more minimized domain gap. Utilizing the cross attention for tokens after self-attention, \cite{ref_zse} introduced a patch-aligning method that can learn the patch-level correspondence between a query sketch and an image candidate. However, calculating the cross-attention on sketch-image may cause tokens with weaker semantic discriminability, resulting miss match during training.

\subsubsection{Language-assisted Vision Models}
Actually,the main challenge of vision-language retrieval is the semantic divergence of heterogeneous data\cite{ref_coor}. Moreover, linguistic information normally contains knowledge that is complementary to image modality. Large Language Models (LLMs), such as GPT-3\cite{ref_lang}, are trained on web-scale datasets. Furthermore, trained LLMs can show impressive capabilities toward few-shot or even zero-shot inference on multiple tasks. Recent papers \cite{ref_i2v,ref_doubly,ref_cascade} have shown that using the textual information generated by LLMs with appropriate prompts can help the model learn complementary knowledge for vision tasks \cite{ref_doubly,ref_cascade}, even for more fine-grained correspondence\cite{ref_i2v}. In this paper, we leverage textual information to align the local patches of sketch and image.

\begin{figure*}[h]
    \centering
    \includegraphics[width=12cm]{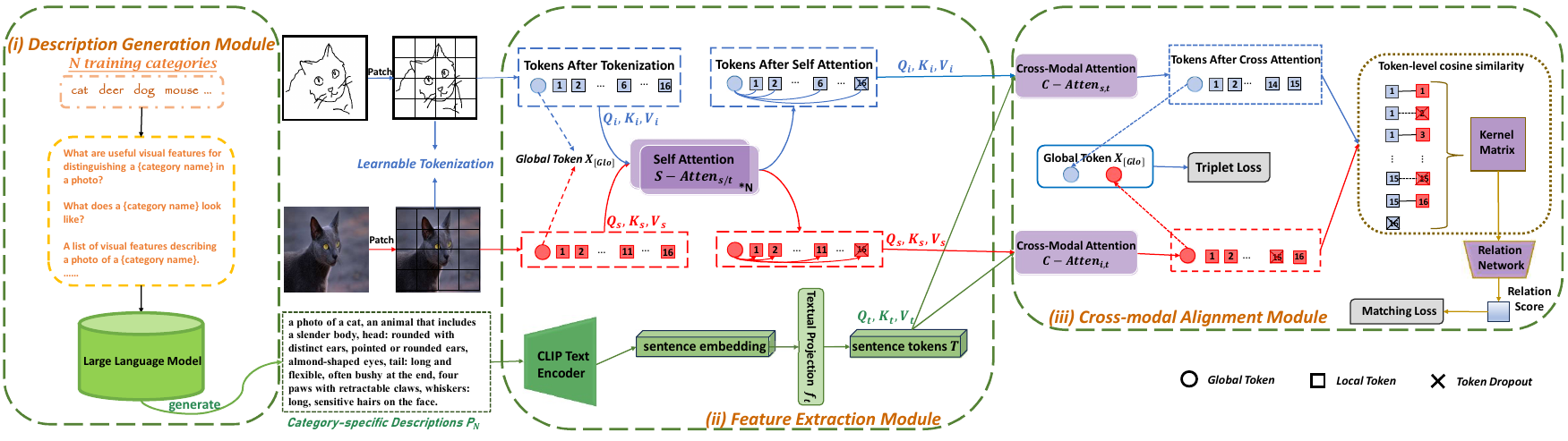}
    \centering
    \caption{Overview. (i) Description Generation Module collects the textual knowledge describing the visual clues for each specific category, which is generated by LLM with some prompts. (ii) Feature Extraction Module takes the sketch, image, and text data and feeds them into the encoders respectively for extracting token-level features by self-attention. (iii) Cross-attention Alignment Module utilizes the cross-attention mechanism to exchange information between sketch-text and image-text tokens. The correspondences of tokens will be measured locally (matching loss) and globally (triplet loss).}
    \label{fig:overview}
\end{figure*}

\section{Proposed Method}
The overall scheme of our proposed framework is shown in Fig.\ref{fig:overview}. Each key module is detailed in the following.
\subsection{Description Generation Module}
Since we want to utilize the zero-shot generalization capability of text information, we need to obtain category-specific text descriptions describing the visual features for training categories. As shown in Fig. \ref{fig:overview}, for \textit{N} training categories, we leverage LLM to produce textual descriptions. For \textit{N} training categories, we can prompt the language model with the inputs: \textit{¡°What are useful visual features for distinguishing a \{category name\} in a photo?"} \cite{ref_visual}(Notably, other appropriate prompts can also be used, which we will discuss in Section 4.3) After prompting, we gather and select numerous sentences describing the distinguishing features of a specific training category. We donate the generated text prompts for \textit{N} categories as $\mathit{P_{N}}$, formulated as:
    \begin{equation}
    \mathit{P_{N}}=\mathrm{LLM}(\mathrm{prompts} ).
\end{equation}

The textual descriptions $\mathit{P_{N}}$ then serve as the input of the text encoder.

\subsection{Feature Extraction Module}

For a given query sketch $S \in \mathbb{R}^{\textit{h}\times\textit{w}\times\textit{c}}$ and a gallery image $I \in \mathbb{R}^{\textit{h}\times\textit{w}\times\textit{c}}$, we use ViT\cite{ref_16} to extract features. More specifically, the ViT partition the input image into non-overlapping patches, then use a projection head to map them into vanilla tokens $S_v\in\mathbb{R}^{\textit{h}\times\textit{w}\times\textit{c}}$
 and image feature $I_v\in\mathbb{R}^{\textit{h}\times\textit{w}\times\textit{c}}$. In addition, the hand-drawn sketch usually contains only the object contour that indicates the primary structure. So, during the feature extraction phase, we adopt the learnable tokenization method to enlarge the receptive field when constructing visual tokens through hierarchical convolution, thereby better preserving structural cues from nearby regions\cite{ref_zse}. It is a network containing four convolution layers and non-linear activation functions. In this way, we can obtain new tokens \textit{$S_n$} and \textit{$I_n$} to adjust the vanilla visual tokens using a residual connection. The final output embedding is formulated as:

\begin{equation}
      \setlength{\abovedisplayskip}{0pt}
     \mathit{X}=\mathit{X_v}+\mathit{X_n},
     \setlength{\belowdisplayskip}{0pt}
\end{equation}
where $\mathit{X}$ stands for Sketch($\mathit{S}$) or Image($\mathit{I}$).

Then, the output embedding serves as the input of ViT. However, we don't use the MLP head offered by the vanilla ViT\cite{ref_16}. We use the feature before the final MLP, which is semantically richer. We call this feature a global token $\mathit{[X_{[Glo]}]}$. As a consequence, the additional learnable global token $ \mathit{[X_{[Glo]}]} \in \mathbb{R}^{d}$ is inserted into the multi-head self-attention and MLP blocks to learn the global representation of an image or a sketch. For both sketch and image, the global tokens after multi-head attention can be attained by:
\begin{equation}
     \mathit{z_{0}} = [\mathit{X_{[Glo]}}; \mathit{X^1};...;\mathit{X^n}],
\end{equation}
\begin{equation}
     \mathit{z_{\ell}} = MSA(LN(z_{\ell-1}))+\mathit{z_{\ell-1}}, \mathit{l}=1...L,
\end{equation}
\begin{equation}
    \mathit{z_{\ell}} = MLP(LN(z_{\ell}))+\mathit{z_{\ell}}, \mathit{l}=1...L,
\end{equation}
where $ \mathit{X} \in \{\mathit{S},\mathit{I}\}$,
LN represents the layer norm. L is the number of layers. And the residual connection is implemented. MSA is the multi-head self-attention, which is formulated as:
\begin{equation}
    \mathit{S-Atten} (Q^{x},K^{x},V^{x}) =  softmax(\frac{Q^{x}{K^{x}}^T}{\sqrt{d} } )V^{x},
\end{equation}
where the $\mathit{Q},\mathit{K},\mathit{V}$ are query, key, value obtained by mapping the same token with three different linear projection heads $[\mathit{{W_{q}}}, \mathit{{W_{k}}}, \mathit{{W_{v}}}]$.

For a given text description generated by LLM, we use the Transformer model to extract its token features. The textual feature $\mathit{T}$ can be attained by:
\begin{equation}
    \mathit{T} = \mathit{f_{t}}(\mathit{E_{t}}(P_{N})),
\end{equation}
where the $\mathit{E_{t}}$ stands for the text encoder, which is the CLIP \cite{ref_learning} text encoder corresponding to the ViT-B/16. $\mathit{f_{t}}$ is the linear layer for projecting the textual tokens to that of the same dimension of image and sketch features.

\subsection{Cross-modal Alignment Module}
Given the features of three different modalities, we aim to align the tokens using cross-attention mechanism.
More specifically, we use text tokens to better align the sketch and image. Cross-modal attention is calculated in sketch-text and image-text. The words describing the same region between an image and a sketch should have larger digits. We update the sketch and image tokens by querying the textual tokens. The tokens after cross-modal attention can be formulated as:
\begin{equation}
    \mathit{C-Atten_{x,t}}  (\mathit{Q_{t}, K_{x}, V_{x}} ) = softmax(\frac{Q_{t}K_{x}^T}{\sqrt{d} } )V_{x}.
\end{equation}

Next, we align the tokens after cross-modal attention. We denote the sketch and image tokens after cross-modal attention as $\mathit{X_{s-t}}$ and $\mathit{X_{i-t}}$. We implement the triplet loss for global alignment. Given a triplet
\textless
$\mathit{{X_{[Glo]}(X_{s-t})}}$, $\mathit{{X_{[Glo]}^{+}(X_{i-t})}}$, $\mathit{{X_{[Glo]}^{-}(X_{i-t})}}$
\textgreater
, where $\mathit{{X_{[Glo]}(X_{s-t})}}$ is the anchor sketch feature, $\mathit{{X_{[Glo]}^{+}(X_{i-t})}}$ is the image feature with the same class label, $\mathit{{X_{[Glo]}^{-}(X_{i-t})}}$ is the image feature with a different class label. The triplet loss pulls the positive pair away from the negative one. Therefore, the triplet loss is formulated as:
\begin{equation}
\begin{aligned}
        \mathit{L_{tri}} = \frac{1}{T} \sum_{i=1}^{T} max\{ \left \| X_{[Glo]}(X_{s-t})-X_{[Glo]}^{+}(X_{i-t}) \right \| -
        \\ \left \| X_{[Glo]}(X_{s-t})-X_{[Glo]}^{-}(X_{i-t}) \right \| +m,0  \} ,
\end{aligned}
\end{equation}
where $\| \cdot \|$  is the L-2 distance, $\mathit{m}$ is the margin parameter, and T donates the total number of triplets.

In addition to aligning the global alignment, features of local patches also need to be handled. We exploit the relation network\cite{ref_zse} to explicitly show the similarity between each pair of visual tokens after cross-modal attention. We first exploit the cosine kernel function, calculating the cosine similarity between all pairs of tokens, resulting in a kernel matrix. Next, a relation network is utilized to explicitly measure the alignment scores. The relation network $\rho$ is a stack of two FC-ReLU-Dropout layers, and the relation score $\mathit{r}$ is a digit in the range of (0,1). The matching score is defined as mean square error(MSE). The relation score $\mathit{r}$ and the matching loss $\mathit{L_{rn}}$ are formulated as:
    \begin{equation}
    \mathit{r}(X_{s-t},X_{i-t})=sigmoid(\rho (\frac{X_{s-t}\cdot X_{i-t}}{\left \| X_{s-t} \right \|\left \| X_{i-t} \right \|  } )),
\end{equation}
\begin{equation}
     \mathit{L_{rn}}=\sum_{i=1}^{N}\sum_{j=1}^{M}(\mathit{r_{i,j}}-pre(y_{i},y_{j}))^2,
\end{equation}
\begin{equation}
     pre = \mathbf{1}(y_{i} == y_{j})
\end{equation}
where $\mathit{pre}$ refers to the score between the predicted score and the ground truth. $r=1$ when matched. y is the class label. N and M are the totals of query sketches and candidate images.

The overall loss $\mathit{L}$ is the summary of $\mathit{L_{tri}}$ and $\mathit{L_{rn}}$:
    \begin{equation}
    \mathit{L} = \lambda_{tri}\mathit{L_{tri}}+\lambda_{rn}\mathit{L_{rn}},
\end{equation}
where $\lambda_{tri}$ and $\lambda_{rn}$ are the hyper-parameters.

\subsection{Inference}
Notably, during inference, we don't use the textual information, only sketch and image features are required. The self attention is implemented on both the sketch and image to produce the uni-modal features. Then, the cross attention is implemented on the uni-modal features to calculate the global and local tokens. Finally, the relation network is used to produce the relation score, which presents the similarity between a query sketch and an image.

\section{Experiment}
\subsection{Implementation}
\textbf{Datasets.} There are three datasets commonly used for ZS-SBIR. \textit{Sketchy}\cite{ref_sketchy} is composed of 75,471 sketches and 12,500 natural images from 125 classes. Sketchy-Ext\cite{ref_fast} is an extended version of the original Sketchy dataset, which contains 125 categories. Moreover, an additional 60,502 photos are included, creating a larger photo gallery. Sketchy-25 refers to a partition of 100 training classes and 25 testing classes. Sketchy-21 \cite{ref_frame} refers to the version of 104/21 train/test classes, which selects classes that do not overlap with ImageNet\cite{ref_imagenet} categories as unseen classes. \textit{TU-Berlin}\cite{ref_tuberlin} contains a total of 250 categories, each category having 80 sketches. It is extended by the collection of 204,489 images provided by \cite{ref_fast}. For the extended TU-Berlin dataset, following the \cite{ref_semantic} we use 220 classes for training and test the remaining 30 classes. \textit{QuickDraw}\cite{ref_doodle} is the largest SBIR dataset with a total of 110 categories. It contains 330,000 sketches drawn by amateurs and 204,000 photos. The split of 80 classes for training and 30 for testing.

\begin{table*}[h]
\begin{center}
     \caption{Comparison results. ¡°-": not reported, ¡°\textdagger": approximate result. The best and second-best scores are colored in \textcolor{red}{red} and \textcolor{blue}{blue}.}
    \centering
    \resizebox{\textwidth}{!}{
    \begin{tabular}{l r r r r r r r r r}
    \toprule[1pt]
    \multirow{2}[4]{*}{\textbf{Methods}} & \multicolumn{1}{l}{\multirow{2}[4]{*}{$\mathit{R^D}$}} & \multicolumn{2}{c}{\textbf{TU-Berlin}} & \multicolumn{2}{c}{\textbf{Sketchy-25}} & \multicolumn{2}{c}{\textbf{Sketchy-21}} & \multicolumn{2}{c}{\textbf{QuickDraw}} \\
\cmidrule{3-10}          &       & \multicolumn{1}{c}{mAP@all} & \multicolumn{1}{l}{Prec@100} & \multicolumn{1}{c}{mAP@all} & \multicolumn{1}{l}{Prec@100} & \multicolumn{1}{c}{mAP@200} & \multicolumn{1}{l}{Prec@200} & \multicolumn{1}{c}{mAP@all} & \multicolumn{1}{l}{Prec@200} \\
    \midrule
    ZSIH\cite{ref_hashing}  & 64    & 0.220  & 0.291  & 0.254  & 0.340  & \--  & \-- & \-- & \-- \\
    CC-DC\cite{ref_gene} & 256   & 0.247  & 0.392  & 0.311  & 0.468  & \-- & \-- & \-- & \-- \\
    DOODLE\cite{ref_doodle} & 256   & 0.109  & \-- & 0.369  & \-- & \-- & \-- & 0.075  & 0.068  \\
    SEM-PCYC\cite{ref_tied} & 64    & 0.297  & 0.426  & 0.349  & 0.463  & \-- & \-- & \-- & \-- \\
    SAKE\cite{ref_semantic}  & 512   & 0.475  & 0.599  & 0.547  & 0.692  & 0.497  & 0.598  & 0.130  & 0.179  \\
    SketchGCN\cite{ref_graph} & 300   & 0.323  & 0.505  & 0.382  & 0.538  & \-- & \-- & \-- & \-- \\
    StyleGuide\cite{ref_style} & 200   & 0.254  & 0.355  & 0.376  & 0.484  & 0.358  & 0.400  & \-- & \-- \\
    PDFD\cite{ref_progress}  & 512   & 0.483  & 0.600  & 0.661  & 0.781  & \-- & \-- & \-- & \-- \\
    ViT-Vis\cite{ref_16} & 512   & 0.360  & 0.503  & 0.410  & 0.569  & 0.403  & 0.512  & 0.101  & 0.113  \\
    ViT-Rec\cite{ref_16} & 512   & 0.438  & 0.578  & 0.483  & 0.637  & 0.416  & 0.522  & 0.115  & 0.127  \\
    DSN\cite{ref_domain}   & 512   & 0.484  & 0.591  & 0.583  & 0.704  & \-- & \-- & \-- & \-- \\
    SBTKNet\cite{ref_eff} & 512   & 0.480  & 0.608  & 0.553  & 0.698  & 0.502  & 0.596  & \-- & \-- \\
    Sketch3T\cite{ref_sk3t} & 512   & 0.507  & \-- & 0.575  & \-- & \-- & \-- & \-- & \-- \\
    TVT\cite{ref_tvt}   & 384   & 0.484  & \textcolor{blue}{0.662}  & 0.648  & \textcolor{blue}{0.796}  & \textcolor{red}{0.531}  & 0.618  & \textcolor{red}{0.149}  & \textcolor{red}{0.293}  \\
    ZSE-SBIR-RN\cite{ref_zse} & 512   & \textcolor{blue}{0.540}  & 0.647  & \textcolor{blue}{0.698}  & 0.789  & 0.525  & \textcolor{blue}{0.620}  & \textcolor{blue}{0.145}  & \textcolor{blue}{0.216}  \\
    \midrule
    Ours  & 512   & \textcolor{red}{0.560}  & \textcolor{red}{0.665}  & \textcolor{red}{0.730}  & \textcolor{red}{0.809}  & \textcolor{blue}{0.528}  & \textcolor{red}{0.624}  & \textdagger{0.139}   & \textdagger{0.209}  \\
    \bottomrule[1pt]
    \end{tabular}
    }
    \label{tab:result}
\end{center}
\end{table*}

\textbf{Competitors.} We compare our model with ZSIH\cite{ref_hashing}, CC-DG\cite{ref_gene},
DOODLE\cite{ref_doodle}, SEM-PCYC\cite{ref_tied}, SAKE\cite{ref_semantic}, SketchGCN\cite{ref_graph}, StyleGuide\cite{ref_style}, PDFD\cite{ref_progress}, ViT-Ret/ViT-Vis\cite{ref_16}, DSN\cite{ref_domain}, SBTKNet\cite{ref_eff}, Sketch3T\cite{ref_sk3t}, TVT\cite{ref_tvt} and ZSE-SBIR\cite{ref_zse}. ViT-Ret: replacing the class token in ViT with a retrieval token. ViT-Vis: using the visual tokens. ZSE-SBIR-RN refers to using the relation network for retrieval.

\textbf{Evaluation protocol.} Following the standard evaluation protocol, we test our model with mean average precision (mAP) and precision on top 100/200 (Prec@100/200).

\textbf{Implementation details.} We implement our model with PyTorch toolkit. A sketch or image is scaled to 224*224. For network architecture, the self-attention consists of 12 layers, ViT-B/16 is pre-trained on ImageNet-1K. The cross-attention is designed as one layer with 12 heads. For the optimizer, we choose Adam with weight decay 1e-2 and learning rate 5e-6.

\subsection{Results}
From Table \ref{tab:result}, we can see that our proposed method achieves all top-3 results over all competitors. More specifically, our method achieves the best result on the Sketchy-25 and TU-Berlin datasets. The reason why our model didn't get a top-1 ranking on Sketchy-21 may be the unseen classes don't overlap with ImageNet categories that ViT-B/16 is trained on. Due to the large data offered by QuickDraw, we test our method by randomly sampling 40000 sketches to retrieve images. The TVT\cite{ref_tvt} got the best result on QuickDraw, maybe because it has (i) a Fusion ViT for distilling tokens of different modalities, (ii) a classification loss, which is capable of learning more zero-shot generalization knowledge on a large-scale noisy data. (iii) the quality of sketches in QuickDraw dataset is not very good, because our method focus more on details, the noise of sketches may cause problem to our method. However, our result, where the \textdagger{} is set, is still comparable.

\begin{table}[h]
  \centering
  \small
  \caption{Ablation study results on manifesting importance of the usage of textual information.}
  \resizebox{8.3 cm}{!}{
    \begin{tabular}{lrrrr}
    \toprule
    \multicolumn{1}{c}{\multirow{2}[4]{*}{Methods}} & \multicolumn{2}{c}{Sketchy} & \multicolumn{2}{c}{TU-Berlin} \\
\cmidrule{2-5}          & \multicolumn{1}{l}{mAP@all} & \multicolumn{1}{l}{Prec@100} & \multicolumn{1}{l}{mAP@all} & \multicolumn{1}{l}{Prec@100} \\    \midrule
    w/o T-S &     0.686    &   0.788    &  0.536     & 0.631 \\
    w/o T-I &     0.689    &    0.790   &  0.556     & 0.649 \\
    w/o text &    0.698    &   0.789    &  0.540     & 0.647 \\
    \midrule
    \rowcolor{gray!30}{Ours}  &   0.730   &   0.809    &    0.560   & 0.665 \\
    \bottomrule[1pt]
    \end{tabular}
  }
  \label{tab:ablation}
\end{table}

\subsection{Further Study}
\textbf{Ablation study.} We aim to test the effectiveness of using textual information in various ways. It's worth noticing that the descriptions yielded by the LLM are only used on training phase. \textbf{w/o T-S}: The cross attention between the text tokens and the sketch tokens are removed, leaving the rest of the network untouched.  \textbf{w/o T-I}: The cross attention between the text tokens and the image tokens are removed, leaving the rest of the network untouched. \textbf{w/o text}: The textual information is completely removed, directly calculating the cross attention between sketch and image tokens.  We can see from Table \ref{tab:ablation}: (i) Without the cross-attention of both text-sketch and image-text, the performance drops significantly, indicating the importance of using text information. (ii) The performance of using only sketch-text cross-attention is better than using only image-text cross-attention, which is maybe because the background of images is more complex compared to sketches.

\textbf{Test of text prompt variation.} We test the effect of utilizing different text prompts on GPT-3. Text prompt1 is not generated by the LLM. It's the fixed prompt "a photo of a \{category name\}." For the remaining ones, we select and collect useful outputs as our model's input. The prompts are inspired by previous works and modified by very little changes. (We only tried those three versions of prompt, other appropriate prompts may also work.) Text prompts 1 to 4 are:

\begin{compactitem}
\item a photo of a \{category name\}.
\item A caption describing a photo of a \{category name\}.~\cite{ref_cascade}
\item What does a \{category name\} look like?~\cite{ref_cascade}
\item What are useful visual features for distinguishing a \{category name\} in a photo?~\cite{ref_visual}
\end{compactitem}

As shown in Table. \ref{tab:text}, the result shows that using text prompt4 got the best output. The simple fixes version of texts brings little benefit to the performance, indicating the importance of text generated from LLM, which describes the attributions of categories in detail. Because by prompting the LLM with text prompt4 can offer more detailed information for recognizing an image, making it possible for us to gather descriptions summarizing the shared attributes of both sketches and images.  It demonstrates that the more specific prompt can lead to better description generated by the LLM, thus boosting model's performance.

\begin{table}[h]
  \centering
  \small
  \caption{Results on Sketchy-25 and Sketchy-21 datasets using texts generated by different text prompts on GPT-3.}
    \begin{tabular}{lrrrr}
    \toprule
    \multicolumn{1}{c}{\multirow{2}[4]{*}{Prompt}} & \multicolumn{2}{c}{Sketchy-25} & \multicolumn{2}{c}{Sketchy-21} \\
\cmidrule{2-5}          & \multicolumn{1}{l}{mAP@all} & \multicolumn{1}{l}{Prec@100} & \multicolumn{1}{l}{mAP@200} & \multicolumn{1}{l}{Prec@200} \\
    \midrule
    Text prompt1 &     0.695    &   0.796    &  0.522     & 0.621 \\
    Text prompt2 &     0.720    &   0.801    &  0.520     & 0.618 \\
    Text prompt3 &     0.727    &   0.807    &  0.525     & 0.618 \\
    \rowcolor{gray!30}{Text prompt4} &     0.730    &   0.809    &  0.528     & 0.624 \\
    \bottomrule[1pt]
    \end{tabular}
  \label{tab:text}
\end{table}

\begin{figure}[h]
    \centering
    \includegraphics[width=12cm]{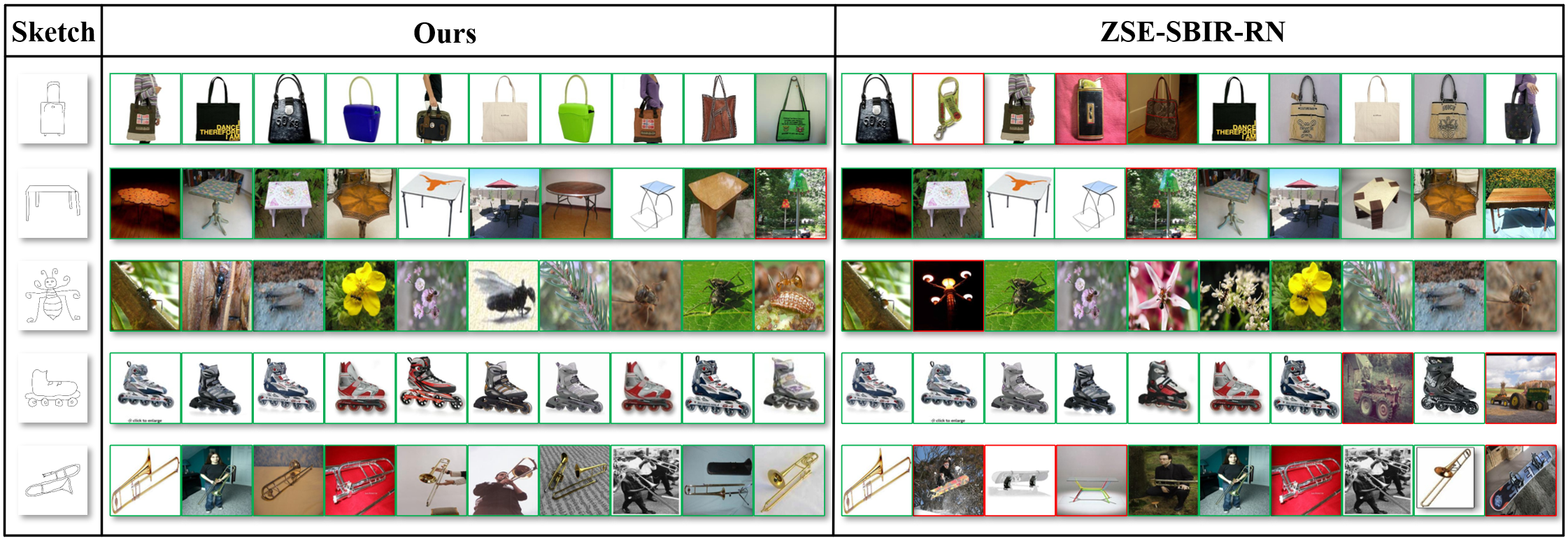}
    \caption{Exemplar comparison retrieval results for the given query sketches and the top 10 retrieved images. \textcolor{red}{Red} box denotes false positive, \textcolor{green}{Green} box denotes true positive.}
    \label{fig:qua}
\end{figure}

\textbf{Qualitative analysis.} The top 10 retrieved candidates of sketches queries are shown in the Fig. \ref{fig:qua}. We can observe that, compared with ZSE-SBIR-RN, our model retrieves better. ZSE-SBIR-RN retrieves false positive images that share similar overall object pose and shape but may ignore some key features. For example, the shape of a suitcase is similar to that of a lighter, both include a rectangular body, but the sketch query of a suitcase that includes a handle. Our model successfully retrieves the true positive images of a suitcase with a handle, corresponding to the text description of a suitcase "equipped with handles". It shows the effectiveness of our model utilizing the textual information, not just the similar shape shared by the query sketches and image candidates.

\begin{figure}[h]
    \centering
    \includegraphics[width=12cm]{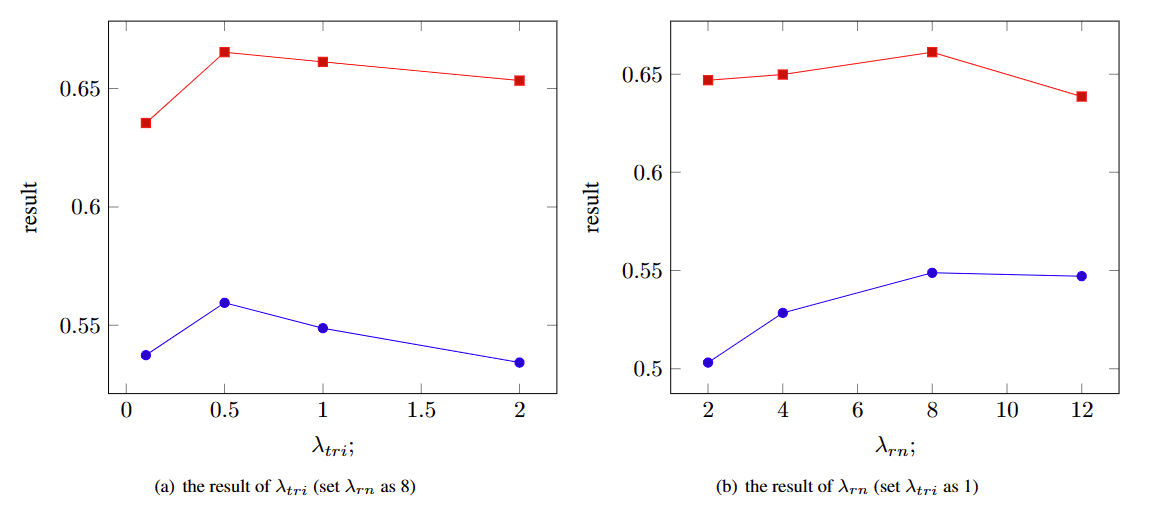}
    \caption{The \textcolor{blue}{mAP@all} and \textcolor{red}{Prec@100}  scores on TU-Berlin with different values of $\lambda_{tri}$ and $\lambda_{rn}$.}
    \label{fig:para}
\end{figure}

\textbf{Analysis on parameter sensitivity.} As shown in Fig. \ref{fig:para}, we analyze the effect of $\lambda_{tri}$ and $\lambda_{rn}$ on TU-Berlin dataset. We can observe that when $\lambda_{rn}$ is set as 8, we take $\lambda_{tri}$'s value of 0.1, 0.5, 1, and 2, the result of both mAP@all and Prec@100 increase from 0.1 to 0.5 and decrease from 0.5 to 2. So, we choose to set $\lambda_{tri}$ as 0.5. Similarly, when $\lambda_{tri}$ is set as 1, we can observe the value of $\lambda_{rn}$ should be set as 8.

\section{Conclusion}
In this work, we introduce an Cross-Modal Attention Alignment Network with Auxiliary Text Description for zero-shot sketch-based image retrieval. The usage of texts in the ZS-SBIR problem demonstrates the rich zero-shot generalization of linguistic data. We gather and select conclusive sentences for each training category by prompting an LLM appropriately. Then, extract feature tokens by self-attention mechanism, exchange the tokens with cross-attention mechanism, and finally align them both locally and globally. Extensive experiments conducted on three benchmark datasets to demonstrate the superiority of our approach. Moreover, we also serve the LLM with different prompts to collect category-specific descriptions for our experiments, comparing the effect of using different prompts.

\section{Acknowledgements}
This work is partially supported by the National Science Foundation of China (62106108, 62276138, 62076135, and 61876087) and the Natural Science Foundation of Jiangsu Province (BK20210559).

%
%
%
%

\end{document}